\documentclass{article}

\usepackage{microtype}
\usepackage{graphicx}
\usepackage{subcaption}
\usepackage{booktabs}
\usepackage{hyperref}

\usepackage[accepted]{icml2026}

\usepackage{amsmath}
\usepackage{amssymb}
\usepackage{mathtools}

\icmltitlerunning{KinEMbed: Decoding Kinematics from Electromyography via Cross-Modal Contrastive Learning}

\begin{document}

\twocolumn[
  \icmltitle{KinEMbed: Decoding Kinematics from Electromyography \\ via Cross-Modal Contrastive Learning}

  \icmlsetsymbol{equal}{*}

  \begin{icmlauthorlist}
    \icmlauthor{Sofia Gilardini}{ox}
    \icmlauthor{Chenfei Ma}{ed}
    \icmlauthor{Kianoush Nazarpour}{ed}
  \end{icmlauthorlist}

  \icmlaffiliation{ox}{Department of Statistics, University of Oxford, Oxford, United Kingdom}
  \icmlaffiliation{ed}{School of Informatics, University of Edinburgh, Edinburgh, United Kingdom}

  \icmlcorrespondingauthor{Sofia Gilardini}{sofia.gilardini@stats.ox.ac.uk}

  \icmlkeywords{contrastive learning, EMG, hand kinematics, regression, wearable sensing, representation learning}

  \vskip 0.3in
]
\printAffiliationsAndNotice{}

% ─────────────────────────────────────────────────────────────────────────────
\begin{abstract}
Decoding hand kinematics from surface electromyography (EMG) is a core
challenge in wearable biosignal processing, underpinning applications across consumer and
commercial human--machine interfaces, virtual and augmented reality, and clinical settings
such as prosthetic control and motor rehabilitation. Most representation learning approaches for EMG
focus on discrete gesture classification, and few focus on continuous regression. We present \textbf{KinEMbed}, a
cross-modal contrastive learning framework for hand kinematics regression that jointly trains dual encoders -- one for
windowed EMG features and one for kinematic (joint angle) targets. The resulting EMG embeddings
are shaped by the geometric structure of the kinematic space without requiring kinematic
signals at inference time.
Evaluating on the NinaPro DB8 dataset that includes both able-bodied subjects and subjects with limb difference ($N{=}11$), KinEMbed attains the highest mean $R^2$ among PCA, PLS, autoencoder and contrastive (CEBRA) baselines on held-out sessions, with the clearest gains on the most challenging thumb degrees of actuation. We
position this work as a first step toward contrastive representation learning for regression of hand kinematics from
structured wearable biosignals. % and a foundation for future work in cross-subject adaptation and pre-training. 

% Decoding hand kinematics from surface electromyography (sEMG) is a core
% challenge in wearable biosignal processing with clinical relevance for prosthetic control
% and motor rehabilitation. Most representation learning approaches for sEMG
% focus on discrete gesture classification, and few focus on continuous regression. We present \textbf{KinEMbed}, a
% cross-modal contrastive learning framework for hand kinematics regression that jointly trains dual encoders -- one for
% windowed sEMG features and one for kinematic (joint angle) targets. The resulting embeddings
% inherit the geometric structure of the kinematic space without requiring kinematic
% signals at inference time. 
% We use the NinaPro DB8 dataset that includes both able-bodied users and subjects with limb difference ($N{=}11$). KinEMbed outperforms PCA, PLS, autoencoder and contrastive (CEBRA) baselines on held-out sessions,
% with largest gains on the most challenging thumb degrees of articulation. We
% position this work as a first step toward contrastive representation learning for regression of hand kinematics from
% structured wearable biosignals. % and a foundation for future work in cross-subject adaptation and pre-training. 
\end{abstract}

% ─────────────────────────────────────────────────────────────────────────────
\section{Introduction}
\label{sec:intro}

Decoding continuous hand joint kinematics from surface electromyography (EMG)
is a longstanding challenge with direct implications across a range of wearable
human--machine interfaces -- from consumer input devices and virtual and
augmented reality to clinical applications such as prosthetic control and motor
rehabilitation. The clinical setting is especially demanding: despite decades of progress in myoelectric prosthetics, upper-limb prosthesis abandonment rates have remained at
approximately 44\% over the past two decades, with inadequate device control
cited as a primary driver of
rejection~\cite{biddiss2007upper,salminger2022acceptance}. A
key bottleneck is the gap between the discrete, sequential control strategies
deployed in commercial devices~\cite{jiang2012myoelectric} and the continuous,
simultaneous, proportional control required for natural hand use.

% The sEMG decoding literature reflects this gap. The dominant paradigm remains
% discrete gesture
% classification~\cite{farina2014extraction,phinyomark2018feature,raghu2025self,yang2025stcnet},
% in which each EMG window is assigned to one of a finite vocabulary of predefined
% postures. While classification accuracy on benchmark datasets is high, the
% approach is fundamentally limited: it cannot represent the continuous, graded
% nature of hand movement, and the rigid output vocabulary constrains the user
% to a small set of pre-trained grasps. 
The dominant paradigm remains discrete gesture
classification~\cite{farina2014extraction,phinyomark2018feature,raghu2025self,yang2025stcnet},
in which EMG windows are assigned to one of a finite set of predefined
postures -- an approach that cannot represent the continuous, graded nature of
hand movement, limiting user utility.

% Continuous joint angle regression, by
% contrast, offers a path to proportional control across multiple degrees of
% freedom simultaneously. 

% Contrastive representation learning has emerged as a powerful framework for
% learning transferable, geometrically structured embeddings across vision,
% language, and audio, by enforcing alignment between paired data views on a shared
% manifold~\cite{chen2020simple,radford2021clip,oord2018representation}. Applied
% to biosignals, CEBRA~\cite{schneider2023learnable} demonstrated that kinematic
% labels can serve as a sampling oracle to structure single-encoder neural
% embeddings of neural recordings. More recently,
% CPEP~\cite{cui2025cpep} introduced cross-modal contrastive pre-training
% between sEMG and visual hand pose for discrete gesture
% classification. However, no prior work has applied cross-modal contrastive
% alignment to the problem of \emph{continuous} kinematic regression from sEMG, a distinct challenge. % -- a setting in which the contrastive framework faces a qualitatively distinct challenge. 

\begin{figure*}[h]
  \centering
  \includegraphics[width=1.92\columnwidth]{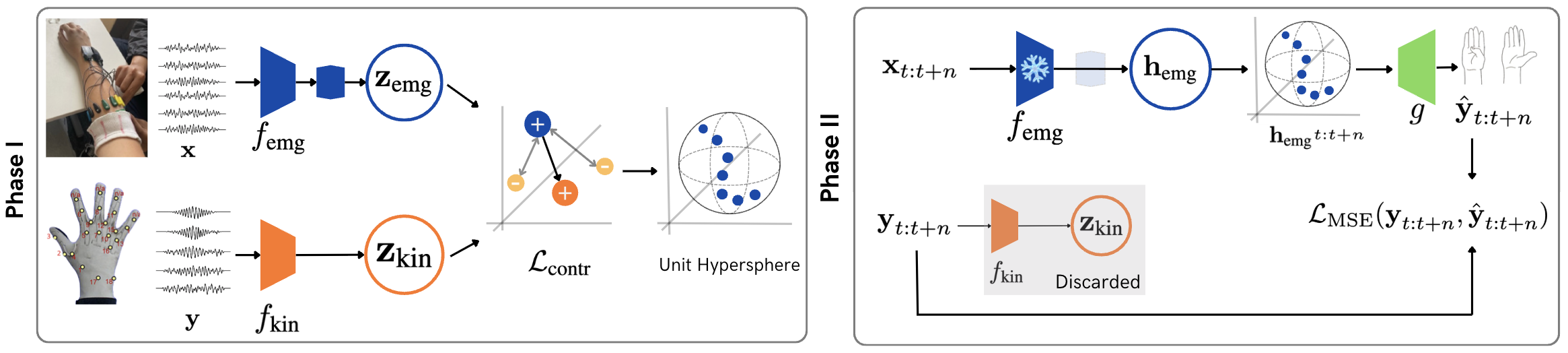}
    \caption{KinEMbed architecture and training phases. In Phase I, two MLP encoders ($f_{\text{emg}}$, $f_\text{kin}$)  project EMG features
           $\mathbf{x}$ and kinematic angles $\mathbf{y}$ onto a shared 16-dimensional unit
           hypersphere trained via a contrastive objective with $\ell_2$-normalisation. In Phase II, the frozen EMG embedding
           is then passed to a TCN decoder ($g$) trained via an MSE objective for continuous DoA regression. Full details in Section~\ref{sec:method}.}
  \label{fig:schematic}
\end{figure*}

Contrastive representation learning is a powerful framework for
learning transferable, structured embeddings within and across
modalities~\cite{chen2020simple,radford2021clip,oord2018representation}. CEBRA~\cite{schneider2023learnable} uses auxiliary variables to guide contrastive sampling for single-encoder neural embeddings.
CPEP~\cite{cui2025cpep} and EMBridge~\cite{cui2026embridge} use cross-modal contrastive pre-training for discrete gesture
classification from EMG. No prior work has applied cross-modal contrastive
alignment to \emph{continuous} kinematic regression from EMG.

We present \textbf{KinEMbed}, the first cross-modal contrastive framework for
continuous EMG-to-kinematics regression. KinEMbed jointly trains dual encoders
-- one for windowed EMG features and one for kinematic targets
-- to align the two modalities on a shared unit hypersphere via a cross-modal InfoNCE loss~\cite{oord2018representation}. The resulting EMG embedding is shaped by the geometric
structure of the kinematic space, providing an inductive bias for downstream regression,
\emph{without} requiring kinematic signals at inference time. We evaluate on the
NinaPro DB8 dataset~\cite{atzori2014electromyography} of $N{=}11$ subjects, including two limb-difference subjects under a strict cross-session protocol in which the test
session is independently recorded and never used during model selection.\\
%Our contributions are summarized as follows: 
%We make the following contributions: 

\vspace{-15pt}

Our main contributions are:

\textbf{1.} We introduce KinEMbed, the first cross-modal contrastive learning framework for continuous hand kinematics regression from EMG, embedding EMG and joint angle vectors as paired modalities in a shared embedding space.

\textbf{2.} We evaluate KinEMbed on a cohort of able-bodied and limb-difference
subjects and benchmark it against five representation-learning baselines (PCA,
PLS, AE-Recon, AE-Super, CEBRA) spanning the supervised/unsupervised and
linear/nonlinear axes, plus ARIMA as a temporal baseline.

\vspace{-2.5mm}

\section{Related Work}
\label{sec:related}

\paragraph{Continuous joint angle regression from EMG.}
While discrete gesture classification has dominated the EMG decoding
literature, a substantial
body of work has pursued continuous joint angle regression for proportional
prosthetic control. Classical approaches apply linear regression, kernel ridge
regression, Gaussian process regression, or support vector regression to
hand-crafted time-domain and frequency-domain feature
windows~\cite{hahne2014linear,muceli2012simultaneous,xiloyannis2017gp}. ~\citet{krasoulis2019effect} evaluated regression-based decoding of finger kinematics on the NinaPro DB8 dataset -- the same dataset used in this work -- and demonstrated accurate per-subject prediction of five degrees of actuation using EMG features and a regularized Wiener filter. More recent deep learning
approaches apply CNNs, LSTMs, and temporal convolutional networks (TCNs)
directly to raw or lightly processed EMG
signals~\cite{zanghieri2019robust}. 

% \paragraph{Contrastive and self-supervised learning for biosignals.}
% SimCLR~\cite{chen2020simple} and InfoNCE~\cite{oord2018representation}
% established that contrastive objectives learn representations that transfer well
% across tasks, with the SimCLR projection head - discarded at inference -
% decoupling the contrastive geometry from the downstream task.
% VICReg~\cite{bardes2022vicreg} extends this to non-contrastive
% variance-invariance-covariance objectives.
% Within the biosignal domain, Cheng et al.~\cite{cheng2020subject} introduced
% subject-aware contrastive learning for EEG and ECG, demonstrating that
% contrastive pre-training with domain-appropriate augmentations produces
% embeddings competitive with fully supervised methods in the low-label regime.
% CEBRA~\cite{schneider2023learnable} uses the InfoNCE loss to align neural
% embeddings with behavioural labels, using kinematics as a sampling oracle to
% define positive \emph{time offsets} rather than encoding kinematics as a
% second modality. Our dual-encoder design is architecturally distinct: we treat
% kinematics as a fully co-equal modality with its own encoder, enforcing explicit
% cross-modal alignment so that the EMG embedding geometry is shaped by the
% \emph{learned} kinematic representation rather than by raw kinematic distances.

\paragraph{Contrastive and self-supervised learning for biosignals.}
SimCLR~\cite{chen2020simple} and InfoNCE~\cite{oord2018representation}
established that contrastive objectives learn representations that transfer well
across tasks, with the SimCLR projection head - discarded at inference -
decoupling the contrastive geometry from the downstream task.
VICReg~\cite{bardes2022vicreg} extends this to non-contrastive
variance-invariance-covariance objectives. CEBRA~\cite{schneider2023learnable} brings contrastive learning to neural
recordings, typically raw and high-dimensional rather than hand-crafted
features. It can be used either \emph{self-supervised} (discovery-driven,
label-free), sampling positive pairs by temporal proximity, or \emph{supervised}
(hypothesis-driven), using an auxiliary variable such as behaviour as a sampling
oracle to define positives; both yield consistent, decodable latent spaces.

% Closely related concurrent work, 
% CPEP~\cite{cui2025cpep} and EMBridge~\cite{cui2026embridge}, apply cross-modal contrastive pre-training for \emph{discrete gesture classification} using visual hand pose data and both visual and EMG data respectively. KinEMbed shares the cross-modal contrastive paradigm but targets \textit{continuous joint angle regression} rather than discrete classification, operates in the per-subject low-data clinical regime including subjects with limb difference, and aligns EMG directly with joint angle vectors rather than visual pose estimates.

% Concurrent work, CPEP~\cite{cui2025cpep} and EMBridge~\cite{cui2026embridge}, use cross-modal contrastive pre-training for \emph{discrete gesture classification}, leveraging visual hand pose (CPEP) and both visual and EMG data (EMBridge). In contrast, KinEMbed targets \textit{continuous joint angle regression}, operates in low-data per-subject clinical settings (including limb difference), and aligns EMG directly with joint angle vectors.

Closely related work, CPEP~\cite{cui2025cpep} and EMBridge~\cite{cui2026embridge}, use cross-modal contrastive pre-training for \emph{discrete gesture classification}, aligning EMG with pose representations, including from large-scale datasets such as emg2pose~\cite{salter2024emg2pose} and kinematics data. In contrast, KinEMbed targets \textit{continuous joint angle regression}, operating directly on fine-grained kinematics in low-data per-subject clinical settings (including limb difference), rather than predicting gesture labels.

\section{Method}
\label{sec:method}

\paragraph{Problem Formulation.}

Given a window of $C{=}16$ surface EMG channels, we extract a feature vector
$\mathbf{x} \in \mathbb{R}^{256}$ (detailed below) and wish to predict five continuous
degrees of actuation (DoA): thumb rotation, thumb flexion, index flexion, middle
flexion, and ring/little flexion, collectively $\mathbf{y} \in \mathbb{R}^5$. Degrees of actuation are aligned with those of the IH2 Azzurra robotic hand~\cite{krasoulis2019effect, prensilia_ih2azzurra_2023}, and are extracted from a hand motion capture glove using a transformation matrix, as described in Appendix~\ref{app_cyberglove_transform}.

\vspace{-10pt}

\paragraph{EMG Feature Extraction}

EMG signals are windowed at 128\,ms (256 samples at 2000\,Hz) with a 52\,ms stride.
Per channel, we compute time-domain features (RMS, Waveform Length, Log Variance,
Zero Crossings, Slope Sign Changes, MAV, Willison Amplitude, Hjorth Mobility and
Complexity), spectral features (Mean Power Frequency, Median Frequency), and STFT
magnitude across five bands. Concatenating over 16 channels yields a 256-dimensional
feature vector. This feature group was selected over three alternative variants by
measuring baseline regression performance on average across all methods under cross-validated session splits.
All features are standardised using per-training-set $z$-score normalisation.

% \subsection{Kinematics Transform}

% In DB8, the hand joint state data is acquired via a Cyberglove II - a motion capture glove that contains 18 joint-angle measurement sensors, distributed as shown in Figure \ref{fig: other-mac/cyberglove_ninapro}. The glove contains two bend sensors on each finger, four abduction sensors, and sensors measuring thumb crossover, palm arch, wrist flexion and wrist abduction. As has been previously done in [REF], a linear transform was applied to the 18 hand joint states to the 5 DoA of the IH2 Azzurra robotic hand \cite{krasoulis2019effect, prensilia_ih2azzurra_2023}. 

\begin{figure*}[t]
  \centering
  \includegraphics[width=\textwidth]{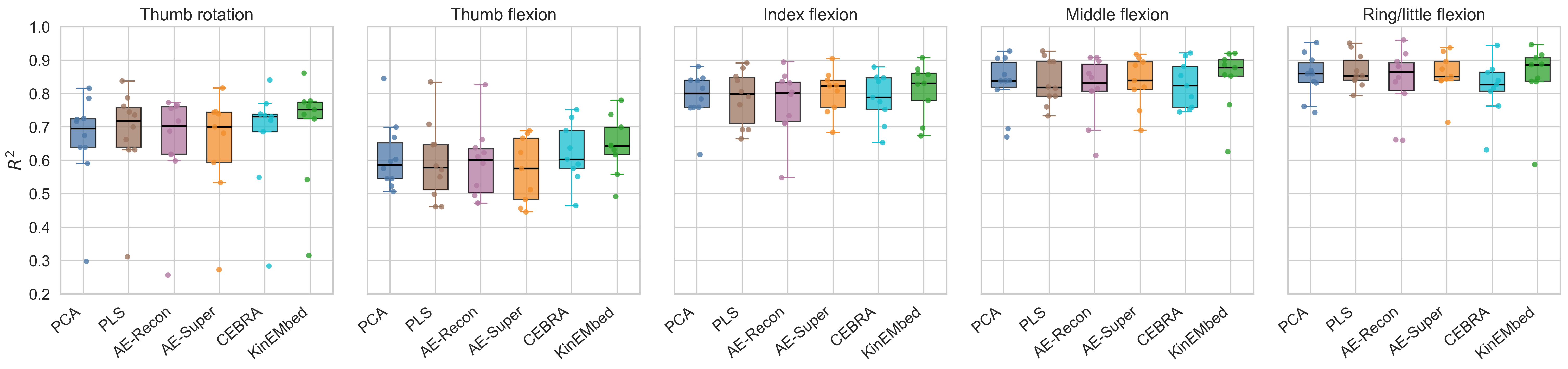}
  \caption{Per-DoA $R^2$ for able-bodied subjects. KinEMbed shows the most
           pronounced gains on thumb rotation and thumb flexion, the most challenging
           and variable degrees of actuation. ARIMA is omitted due to poor performance (negative $R^2)$.}
  \label{fig:perdoa}
  \vspace{-2mm}
\end{figure*}

\subsection{KinEMbed Architecture}

\paragraph{Dual encoders.}
We train two separate MLPs jointly as shown in Figure~\ref{fig:schematic}. The \emph{EMG
encoder} $f_\text{emg}$ has hidden dimensions $[256, 128]$ with batch normalisation,
ReLU activations, and dropout ($p{=}0.1$). The \emph{kinematic encoder} $f_\text{kin}$
has hidden dimensions $[64, 32]$ with the same activations. Both encoders project to
an $E$-dimensional space ($E{=}16$) followed by $\ell_2$-normalisation onto a unit hypersphere:

% \begin{align}
%   \mathbf{z}_\text{emg} &= \ell_2\!\left(f_\text{emg}(\mathbf{x})\right),
%   \quad \mathbf{z}_\text{emg} \in \mathbb{R}^{16}, \\
%   \mathbf{z}_\text{kin} &= \ell_2\!\left(f_\text{kin}(\mathbf{y})\right),
%   \quad \mathbf{z}_\text{kin} \in \mathbb{R}^{16}.
% \end{align}

\begin{align}
  \mathbf{h}_\text{emg} &= \ell_2\!\left(f_\text{emg}(\mathbf{x})\right),
  \quad \mathbf{h}_\text{emg} \in \mathbb{R}^{16}, \\
  \mathbf{z}_\text{kin} &= \ell_2\!\left(f_\text{kin}(\mathbf{y})\right),
  \quad \mathbf{z}_\text{kin} \in \mathbb{R}^{16}.
\end{align}

% \begin{align}
%   \mathbf{z}_\text{emg} &= \ell_2\!\left(f_\text{emg}(\mathbf{x})\right),
%   \quad \mathbf{z}_\text{emg} \in \mathbb{R}^{16}, \\
%   \mathbf{z}_\text{kin} &= \ell_2\!\left(f_\text{kin}(\mathbf{y})\right),
%   \quad \mathbf{z}_\text{kin} \in \mathbb{R}^{16}.
% \end{align}

% \paragraph{Projection head.}
% Following SimCLR~\cite{chen2020simple}, a two-layer nonlinear projection head is
% appended to $f_\text{emg}$ during contrastive training, so the contrastive loss is applied
% to the L2-normalised projection-head output. At inference the projection head is discarded
% and the downstream decoder consumes the L2-normalised encoder trunk
% $\ell_2(f_\text{emg}(\mathbf{x}))$. This decouples the geometry directly optimised by the
% contrastive loss from the geometry seen by the decoder.

\paragraph{Projection head.}
Following SimCLR~\cite{chen2020simple}, a two-layer nonlinear projection head
$g_{proj}:\mathbb{R}^{16}\!\to\!\mathbb{R}^{16}$ (hidden width 32, batch normalisation
and ReLU) is appended to the EMG trunk during contrastive training. The EMG
embedding seen by the contrastive loss is therefore the $\ell_2$-normalised
projection-head output
\begin{equation}
  \mathbf{z}_\text{emg} = \ell_2\!\left(g_{proj}(\mathbf{h}_\text{emg})\right)
  \in \mathbb{R}^{16},
\end{equation}
whereas the kinematic branch has no projection head, so $\mathbf{z}_\text{kin}$
enters the loss directly. At inference the projection head is discarded and the
downstream decoder consumes the encoder representation $h_{emg}$. This decouples the
geometry directly optimised by the contrastive loss (over
$\mathbf{z}_\text{emg}$) from the geometry seen by the decoder (over
$\mathbf{h}_\text{emg}$).

\paragraph{Contrastive objective.}
Given a batch of $N$ synchronised (EMG, kinematic) pairs, we use a cross-modal InfoNCE loss:

\begin{equation}
  \mathcal{L}_\text{contr} = -\frac{1}{2N}\sum_{i=1}^{N}
  \left[
    \log \frac{e^{\mathbf{z}_i^\text{emg} \cdot \mathbf{z}_i^\text{kin}/\tau}}
             {\sum_{j} e^{\mathbf{z}_i^\text{emg} \cdot \mathbf{z}_j^\text{kin}/\tau}}
    +
    \log \frac{e^{\mathbf{z}_i^\text{kin} \cdot \mathbf{z}_i^\text{emg}/\tau}}
             {\sum_{j} e^{\mathbf{z}_i^\text{kin} \cdot \mathbf{z}_j^\text{emg}/\tau}}
  \right],
  \label{eq:ntxent}
\end{equation}

% {\small
% \begin{equation}
%   \mathcal{L} = -\frac{1}{2N}\sum_{i=1}^{N}
%   \left[
%     \log \frac{e^{\mathbf{z}_i^\text{emg} \cdot \mathbf{z}_i^\text{kin}/\tau}}
%              {\sum_{j} e^{\mathbf{z}_i^\text{emg} \cdot \mathbf{z}_j^\text{kin}/\tau}}
%     +
%     \log \frac{e^{\mathbf{z}_i^\text{kin} \cdot \mathbf{z}_i^\text{emg}/\tau}}
%              {\sum_{j} e^{\mathbf{z}_i^\text{kin} \cdot \mathbf{z}_j^\text{emg}/\tau}}
%   \right],
%   \label{eq:ntxent}
% \end{equation}
% }

% \begin{equation}
% \begin{aligned}
% \mathcal{L} = -\frac{1}{2N}\sum_{i=1}^{N} \Bigg[
% &\log \frac{e^{\mathbf{z}_i^\text{emg} \cdot \mathbf{z}_i^\text{kin}/\tau}}
%          {\sum_{j} e^{\mathbf{z}_i^\text{emg} \cdot \mathbf{z}_j^\text{kin}/\tau}} \\
% &+
% \log \frac{e^{\mathbf{z}_i^\text{kin} \cdot \mathbf{z}_i^\text{emg}/\tau}}
%          {\sum_{j} e^{\mathbf{z}_i^\text{kin} \cdot \mathbf{z}_j^\text{emg}/\tau}}
% \Bigg]
% \end{aligned}
% \label{eq:ntxent}
% \end{equation}
where $\tau{=}0.2$ is the temperature and positive pairs are temporally aligned
(EMG window $i$, DoA measurement $i$). We also evaluated a \emph{Soft InfoNCE}
variant that replaces the one-hot contrastive target with a Gaussian-kernel soft
target over DoA distances (addressing the false-negative problem for regression),
and VICReg~\cite{bardes2022vicreg}. The standard InfoNCE loss was selected via grid search, although performance differences were limited
(Section~\ref{sec:experiments}; per-loss results in Appendix~\ref{sec:hparam}).

% \paragraph{Downstream TCN decoder.}
% After contrastive pre-training, the EMG encoder is frozen and a Temporal Convolutional
% Network (TCN)~\cite{bai2018tcn} decoder is trained to map sequences of EMG
% embeddings to 5-dimensional DoA predictions. The TCN ($g$) uses dilated causal convolutions
% (64 channels, kernel size 3, dilations~$[1,2,4]$) with weight normalisation and
% residual connections. The input sequence length $\ell \in \{5,10,20\}$ is selected
% per method per subject via 3-fold CV. The decoder is trained for 100 epochs with
% Adam~\cite{kingma2014adam} ($\eta{=}10^{-3}$, MSE loss). All baselines share
% this decoder, ensuring that embedding quality is the only variable under comparison. Full training details in Appendix~\ref{sec:hparam}, code to be released soon.

\paragraph{Downstream TCN decoder.}
After contrastive pre-training, the EMG encoder is frozen and a Temporal Convolutional
Network (TCN)~\cite{bai2018tcn} decoder is trained to map sequences of EMG
embeddings to 5-dimensional DoA predictions. The TCN ($g$) uses dilated causal convolutions
(64 channels, kernel size 3, dilations~$[1,2,4]$) with weight normalisation and
residual connections. The decoder is trained for 100 epochs with
Adam~\cite{kingma2014adam} ($\eta{=}10^{-3}$, MSE loss). In order to maximise like-for-like comparison, all representation-learning baselines share
this decoder. An MLP, GRU, and TCN were evaluated as potential decoders. TCN was chosen as the best performing on average across all methods, based on the $R^2$ across a subsection of tested subjects via 2-fold CV. Full training details are available in Appendix~\ref{sec:hparam}.
\vspace{-2mm}

% ─────────────────────────────────────────────────────────────────────────────

\begin{table}[t]
  \centering
  \caption{Overall mean $R^2$ (across 5 DoA) on held-out session \texttt{d3}. Values: mean $\pm$ std across subjects ($R^2$). Best per column
           in \textbf{bold}.}
  \label{tab:overall_r2}
  \resizebox{\columnwidth}{!}{%
  \begin{tabular}{lrrrrrr}
    \toprule
    Method & \multicolumn{2}{c}{All ($n$=11)} & \multicolumn{2}{c}{AB ($n$=9)} & \multicolumn{2}{c}{limb difference ($n$=2)} \\
    \midrule
    ARIMA & -0.072 & 0.056 & -0.075 & 0.056 & -0.062 & 0.074 \\
    PCA & 0.717 & 0.113 & 0.748 & 0.077 & 0.560 & 0.170 \\
    PLS & 0.722 & 0.118 & 0.753 & 0.079 & 0.566 & 0.201 \\
    AE-Recon & 0.704 & 0.111 & 0.733 & 0.081 & 0.558 & 0.159 \\
    AE-Super & 0.703 & 0.132 & 0.743 & 0.076 & 0.523 & 0.221 \\
    CEBRA & 0.713 & 0.115 & 0.745 & 0.078 & \textbf{0.570} & 0.181 \\
    \midrule
    \textbf{KinEMbed} & \textbf{0.732} & 0.129 & \textbf{0.769} & 0.090 & 0.568 & 0.191 \\
    \bottomrule
  \end{tabular}
}
\end{table}

\section{Experiments}
\label{sec:experiments}

\paragraph{Dataset and Evaluation Protocol.}

We evaluate on the NinaPro DB8 dataset~\cite{atzori2014electromyography}, which comprises synchronised
EMG and instrumented glove recordings from 12 subjects performing continuous hand
grasps and a wide range of finger movements. Subject 4 was excluded due to low-fidelity data,
yielding $N{=}11$ participants: 9 able-bodied (AB) and 2 LD subjects. Although the LD cohort is small, we deliberately retain it, as evaluating on limb-difference data is essential for advancing continuous, proportional control for prosthetic applications, a central long-term goal of this work.

Each subject completed three sessions (\texttt{d1}, \texttt{d2}, \texttt{d3}). Sessions \texttt{d1} and \texttt{d2} are used for
training; \texttt{d3} is the strictly held-out test set, never touched during model selection.
We use 2-fold session CV during hyperparameter tuning. % (fold 1: train on \texttt{d1}, validate on \texttt{d2}; fold 2: vice versa). 
We report the mean coefficient of determination ($R^2$) across the 5 DoA channels on the held-out session \texttt{d3}. Each subject's $R^2$ is averaged over three seeds, and Table~\ref{tab:overall_r2} reports the mean $\pm$ standard deviation across subjects.

% \subsection{Baselines}

\paragraph{Baseline selection.}
We initially evaluated seven candidate DR methods in preliminary experiments:
PCA, UMAP, PLS, a reconstruction autoencoder (AE-Recon), a supervised autoencoder
(AE-Super), a contrastive autoencoder, a conditional VAE (CVAE) and CEBRA. The five
baselines retained for the final comparison -- PCA, PLS, AE-Recon, AE-Super and CEBRA -- were
selected by ranking all candidates on mean $R^2$ under 2-fold session CV and keeping
the top performers. The set spans the key methodological axes: unsupervised
linear (PCA), supervised linear (PLS), unsupervised nonlinear (AE-Recon), and
supervised nonlinear (AE-Super). \textit{To the best of our knowledge, this is also the first instance of CEBRA applied to EMG decoding.} We also included a non–representation-learning baseline, Autoregressive Integrated Moving Average (ARIMA), to benchmark against a classical statistical time-series model. 

\paragraph{Baseline configurations.}
Because the aim of this work is to \emph{introduce} KinEMbed rather than to produce
an exhaustive benchmark, baseline architectures follow standard designs with
minimal tuning. AE-Recon uses a symmetric MLP encoder-decoder ($256 \to 128 \to 64
\to d$) trained with MSE reconstruction loss (600 epochs, Adam, $\eta{=}10^{-4}$).
AE-Super shares the same encoder but adds an auxiliary DoA regression head from the
bottleneck; the reconstruction/regression loss weighting $\alpha$ is selected from
$\{0.1, 0.5, 1.0\}$ via 2-fold session CV. PCA and PLS are fit on the training split
using standard \texttt{sklearn} implementations. CEBRA was tuned as described in Appendix~\ref{app_cebra_hyperparams}.
To ensure a fair comparison, each baseline's embedding dimension is independently
optimised via grid search over $d \in \{2,3,4,8,16,32\}$ using 2-fold session CV.
The canonical dimension is the mode of per-subject best values (PCA: 32, PLS: 32,
AE-Recon: 32, AE-Super: 16, CEBRA: 8).

\begin{figure}[h]
  \centering
  \includegraphics[width=0.9\columnwidth]{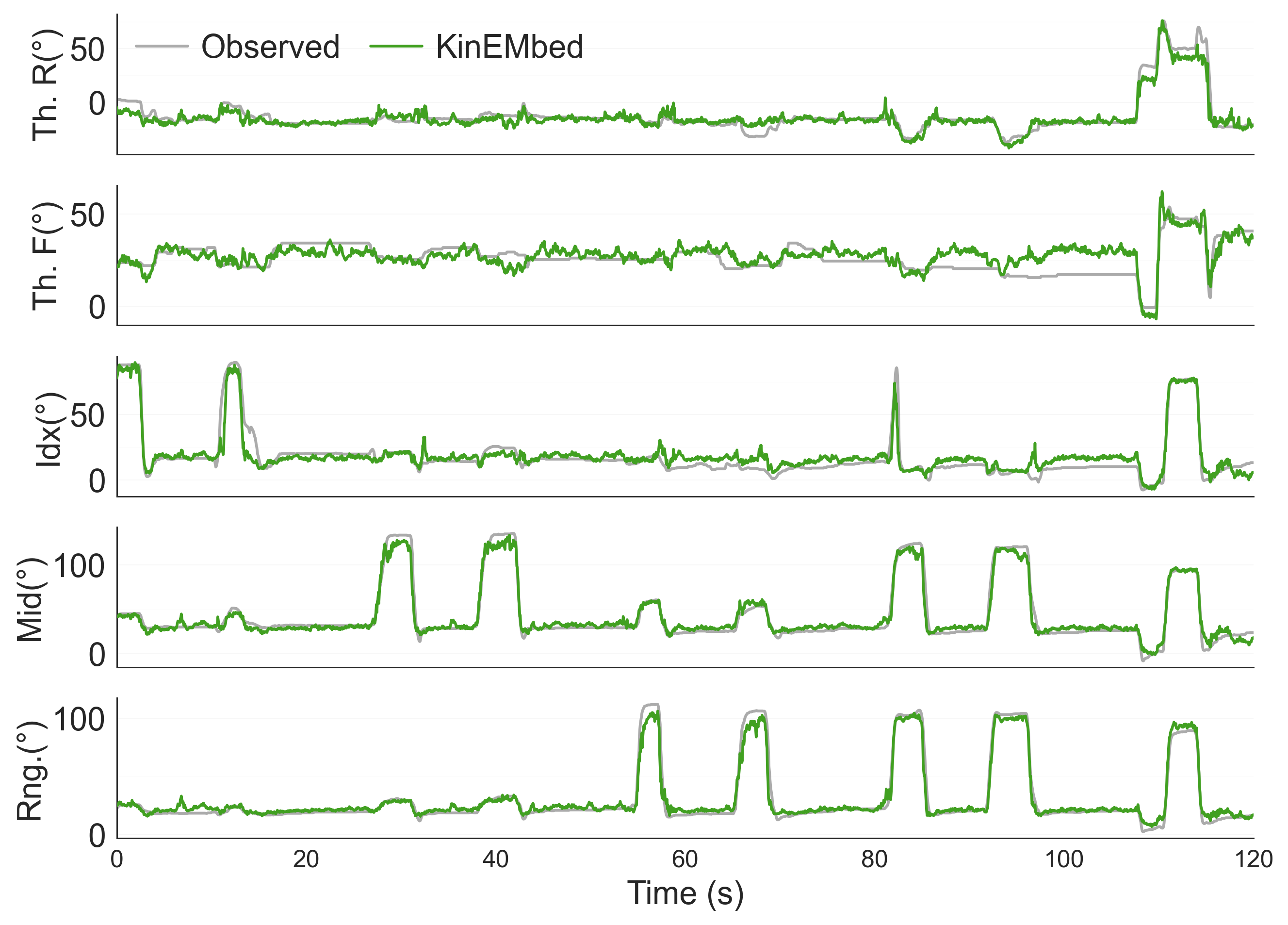}
  \caption{Example 120-second continuous prediction trace for subject 10 (able-bodied).
           All five DoA are decoded simultaneously from EMG using the frozen $f_{emg}$ encoder.}
  \label{fig:trace}
\end{figure}

\subsection{Results}

\begin{figure}[t]
  \centering
  \includegraphics[width=0.9\columnwidth]{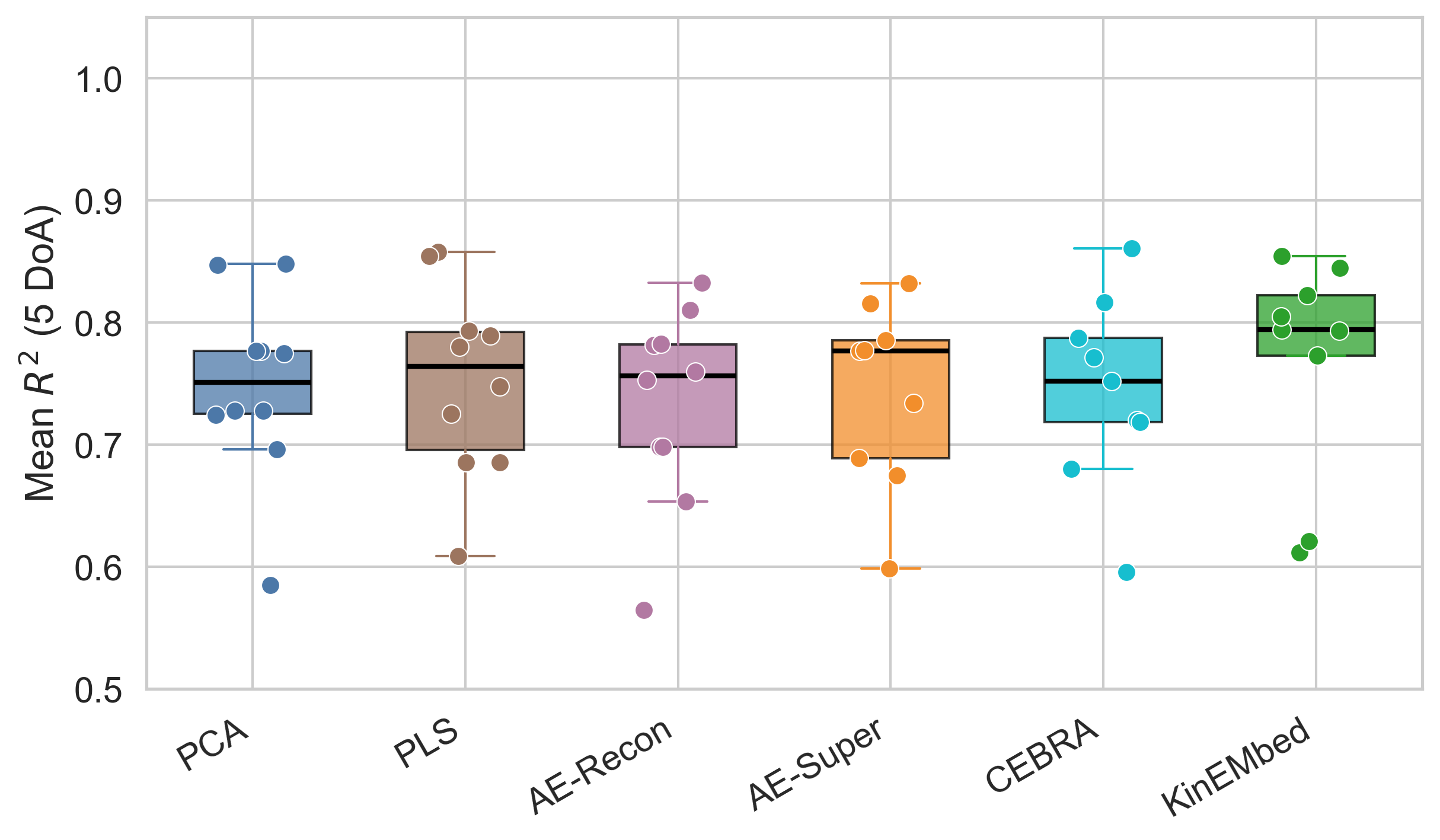}
  \caption{Distribution of mean $R^2$ across able-bodied subjects ($n{=}9$) on held-out
           session d3. KinEMbed achieves the highest median and mean $R^2$.}
  \label{fig:boxplot}

\end{figure}

Table~\ref{tab:overall_r2} reports overall mean $R^2$ across the 5 DoA for all
subjects. KinEMbed achieves the highest mean $R^2$ overall, with a mean of \textbf{0.732} (all subjects), \textbf{0.769}
(AB), and \textbf{0.568} (LD). Figure~\ref{fig:boxplot}
illustrates the $R^2$ distribution across AB subjects. Within the top-performing methods, KinEMbed shows a tighter inter-quartile distribution. ARIMA, which exploits
only the temporal autocorrelation of the DoA signal without any EMG input, achieves
\textit{negative} $R^2$ across all groups, confirming that EMG data carries essential
non-redundant kinematic information. On the LD cohort, KinEMbed and CEBRA attain comparable performance, mean $R^2$ (0.568 vs.\ 0.570, respectively). With only two LD subjects and high inter-subject variance, no firm conclusions on performance can be drawn.

Table~\ref{tab:per_doa_r2} breaks results down by DoA for AB subjects.
KinEMbed achieves the best mean $R^2$ and leads on four of five DoA. The largest
gains are on \emph{thumb flexion} (+2.3 pp over the next-best method, CEBRA) and \emph{thumb rotation}
(+1.4 pp), which are the most variable and clinically relevant degrees. The ring/little
finger DoA is already well-predicted by all methods and differences are negligible, as shown in 
Figure~\ref{fig:perdoa}.

\section{Discussion and Conclusion}
\label{sec:conclusion}

We present KinEMbed as an initial demonstration that cross-modal contrastive alignment is a viable and competitive paradigm for continuous decoding of hand kinematics from EMG, rather than as an exhaustively optimised system. Throughout this work, we prioritised controlled, like-for-like comparison over absolute performance: every representation method is given the same hand-crafted feature input and the same downstream decoder. The decoder exemplifies this trade-off -- we adopt a TCN because it was the best performer \emph{on average across all methods}, a choice that limits confounders but is, by design, not tailored to any individual method, KinEMbed included. Such conservative choices make the reported advantage more naturally read as a lower bound than a ceiling: a feature pipeline, encoder, or decoder optimised specifically for each KinEMbed could plausibly widen the margin. A more exhaustive exploration of these components is a clear and promising next step. 

While the cohort is limited in size, the evaluation is deliberately stringent: every method is assessed under a strict cross-session protocol, training and testing on independently recorded sessions. This is the appropriate standard for both commercial and clinically motivated applications due to distribution shift caused by a variety of factors including fatigue, sensor placement, and changes in the signal-to-noise ratio. This protocol is considerably more demanding than the within-session splits commonly seen in the literature.

\begin{table}[t]
  \centering
  \caption{Per-DoA $R^2$ on held-out session d3, able-bodied subjects ($n{=}9$).
           Best per column in \textbf{bold}. DoA names abbreviated: Th.R = thumb
           rotation, Th.F = thumb flexion, Idx = index, Mid = middle, Rng = ring/little.}
  \label{tab:per_doa_r2}
  \resizebox{\columnwidth}{!}{%
  \begin{tabular}{lrrrrrc}
    \toprule
    Method & Th.R & Th.F & Idx & Mid & Rng & Mean \\
    \midrule
    ARIMA & $-0.045$ & $-0.025$ & $-0.009$ & $-0.008$ & $-0.286$ & $-0.075$ \\
    PCA & $0.660$ & $0.611$ & $0.790$ & $0.828$ & $0.853$ & $0.748$ \\
    PLS & $0.680$ & $0.596$ & $0.787$ & $0.834$ & $\mathbf{0.867}$ & $0.753$ \\
    AE-Recon & $0.655$ & $0.591$ & $0.772$ & $0.815$ & $0.834$ & $0.733$ \\
    AE-Super & $0.647$ & $0.570$ & $0.805$ & $0.834$ & $0.857$ & $0.743$ \\
    CEBRA & $0.672$ & $0.621$ & $0.787$ & $0.826$ & $0.818$ & $0.745$ \\
    \midrule
    \textbf{KinEMbed} & $\mathbf{0.694}$ & $\mathbf{0.644}$ & $\mathbf{0.812}$ & $\mathbf{0.845}$ & $0.850$ & $\mathbf{0.769}$ \\
    \bottomrule
  \end{tabular}
  }
\end{table}

% KinEMbed introduces cross-modal contrastive learning as a new paradigm for continuous
% hand joint regression from EMG. By treating EMG and kinematics as co-equal modalities
% and aligning their latent representations on a shared unit sphere, the contrastive
% objective shapes the shared EMG trunk towards a representation aligned with the kinematic
% space -- an alignment that reconstructive and linear baselines do not exploit. KinEMbed
% attains the highest mean $R^2$ overall, although its margin over the strongest baselines
% (PLS, PCA, CEBRA) is small; with only a small cohort of subjects these
% differences may lie within noise, and we therefore read them as suggestive rather than
% conclusive (see Limitations).

% However, it is promising that the gains are most pronounced for thumb movements, which are mechanically decoupled
% from the other fingers and produce more variable EMG patterns, precisely where
% discriminative contrastive pressure provides the most benefit over reconstructive objectives.

\paragraph{Limitations.}
This work introduces KinEMbed and presents an initial evaluation against relevant
baselines; however several limitations temper its conclusions. \textbf{(i)} Improvements over the
strongest baselines are small in absolute terms and, with only $N{=}11$ subjects, the
study is underpowered to establish statistical significance; we therefore treat the method
rankings as suggestive and defer significance testing on larger cohorts to future work.
\textbf{(ii)} The limb-difference cohort comprises only two subjects, too few to draw firm
conclusions for prosthetic applications. \textbf{(iii)} We do not characterise when the false-negative problem matters:
the Soft InfoNCE variant designed to address it did not outperform the standard InfoNCE in
our limited grid search, leaving open whether the effect is genuinely mild or
simply undetected at this scale.

% \paragraph{Future directions.}
% Several natural extensions follow from this work. \emph{Temporal contrastive objectives}
% (e.g., CPC-style) could exploit the sequential structure of EMG more explicitly.
% \emph{Cross-subject and cross-session adaptation} via contrastive pre-training on pooled
% data, followed by subject-specific fine-tuning, is a direct path to clinical deployment.
% The dual-encoder structure is also well-suited to \emph{foundation model pre-training}:
% a kinematic encoder pre-trained on large motion-capture datasets could provide a rich
% target embedding space without requiring synchronised EMG.

\vspace{-10pt}

\paragraph{Future directions.}
The most natural next step is stronger evaluation across larger and
more diverse cohorts -- in particular more participants with limb difference.
Together with paired significance tests and bootstrap confidence intervals, this would
allow for more conclusive analysis on the performance of this methodology. A complementary direction is \emph{mechanistic analysis}
of the learned representation and a more systematic
study of the contrastive objective --including the conditions under which the
Soft-InfoNCE variant meaningfully mitigates the false-negative problem -- which would
resolve the question left open above. Several methodological extensions also
follow naturally. Similarly, \emph{temporal contrastive objectives} (e.g., CPC-style) could
exploit the sequential structure of EMG more explicitly.

\emph{Cross-subject and cross-session adaptation}, via contrastive pre-training
on pooled data followed by subject-specific fine-tuning, offers a direct path to
clinical deployment with reduced per-user calibration. The dual-encoder design is
also well suited to \emph{foundation-model pre-training}, in which a kinematic
encoder pre-trained on large motion-capture corpora supplies a rich target
embedding space without requiring synchronised EMG. 

We hope KinEMbed serves as a useful baseline for EMG regression tasks and a conceptual stepping stone for representation learning on structured wearable biosignals.\\

\section*{Acknowledgments}
S.G. thanks Dr Sadra Sadeh for his early mentorship on a closely related project, which sparked her interest in this area of research.

\section*{Impact Statement}

This paper presents work whose goal it is to advance the field of Machine Learning. There are many potential societal consequences of our work, none of which we feel must be specifically highlighted here.

\vspace{20pt}

%Add alternative methods (not just EMG). 

% ─────────────────────────────────────────────────────────────────────────────
\bibliography{emg_cmcr_paper}
\bibliographystyle{icml2026}

% ─────────────────────────────────────────────────────────────────────────────
\newpage
\appendix

\section{Dataset}

\begin{figure}[H]
    \centering
    \includegraphics[width = 0.8\columnwidth]{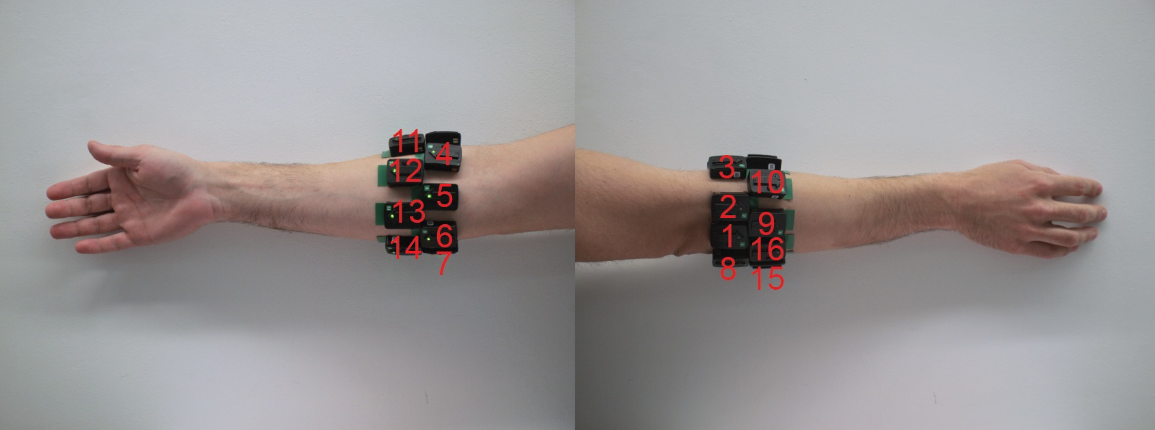}
    \caption{The positioning of the 16 electrodes in the NinaPro dataset. Image reproduced from \cite{krasoulis2019ninapro}.}
    \label{fig: emg_electrodes_ninapro}
\end{figure}

\begin{figure}[h]
    \centering
    \subfloat[]{
        \includegraphics[width=0.45\columnwidth]{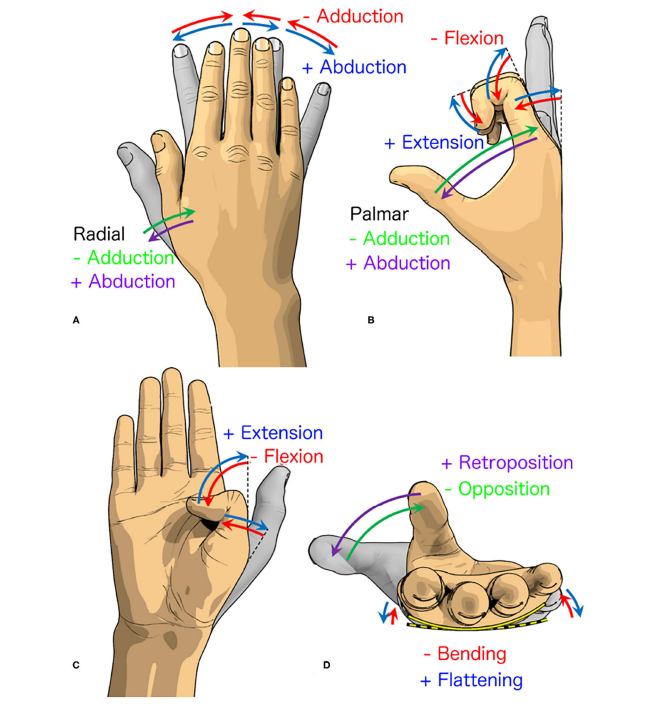}
        \label{fig: ad_ab_diagram_cabibihan2021suitability}
    }
    \hfill
    \subfloat[]{
        \includegraphics[width=0.45\columnwidth]{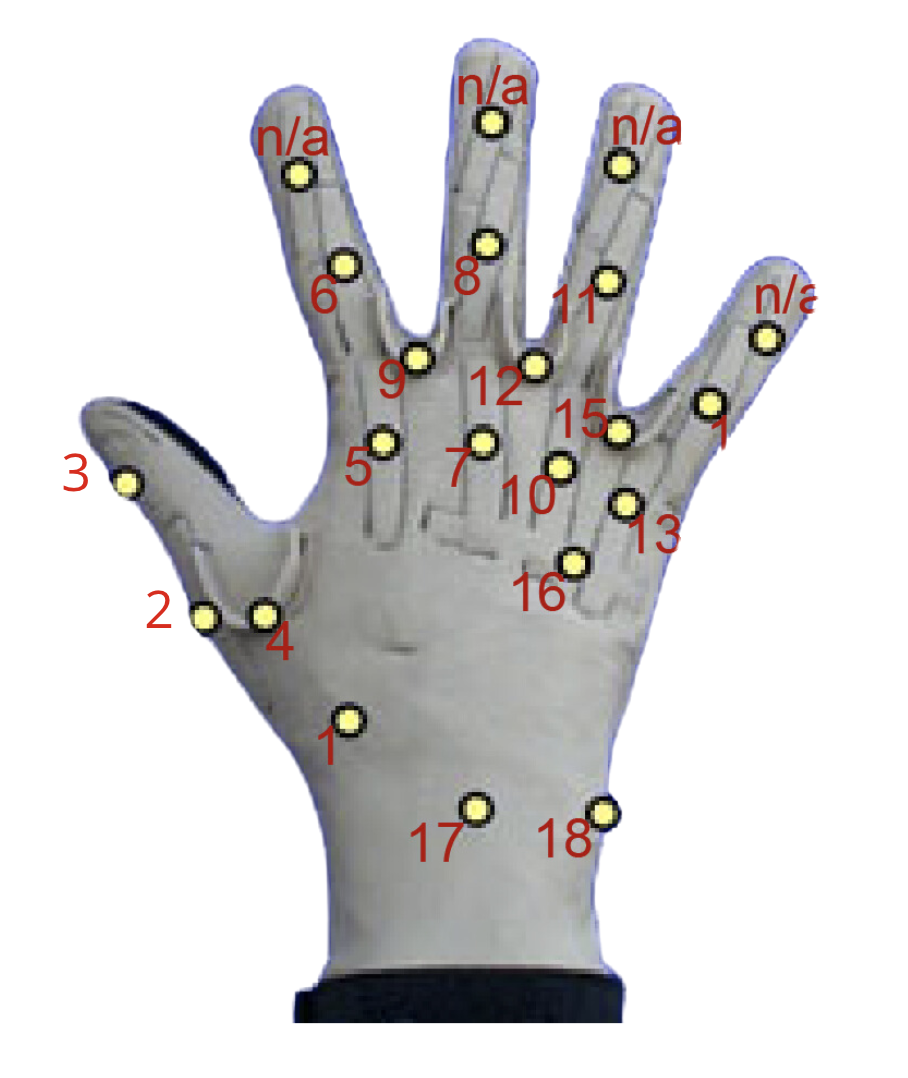}
        \label{fig: dataset/cyberglove_ninapro}
    }
    \caption{A visualisation of the movements of the hand and wrist (a). Image reproduced from \cite{cabibihan2021suitability}. The positioning of the joint state sensors (b). Image reproduced from \cite{krasoulis2019ninapro}.}
\end{figure}

We evaluate KinEMbed and baselines on the NinaPro DB8 dataset. The EMG data in DB8 was recorded using 16 active double-differential wireless sensors from a Delsys Trigno IM Wireless EMG system \cite{6290287} as shown in Figure~\ref{fig: emg_electrodes_ninapro}. Muscle activity was recorded from the participants' right forearm (i.e., the remnant limb for subjects with limb difference). Motion capture data was recorded with a Cyberglove~II - a motion capture glove that contains 18 joint-angle measurement sensors, distributed as shown in Figure~\ref{fig: dataset/cyberglove_ninapro}.

\subsection{Cyberglove Data}
\label{app_cyberglove_transform}

Although the Cyberglove II has 18 joint measurement sensors, the objective of this research is the control of an external device such as a dexterous robotic hand, or a simulated hand. Hence, a mapping strategy is employed to translate the 18 channels of joint angle data to DoA of a robotic hand. The mapping of Cyberglove joint states to hand position reconstruction has been explored for visualisation \cite{wang2013data} and robot manipulation \cite{steffen2011robust} and has further been refined for the actuation of a robotic hand for prostheses using a linear mapping from the 18 joint states to the 5 DoA of the IH2 Azzurra robotic hand \cite{krasoulis2019effect, prensilia_ih2azzurra_2023}. In this transformation, the 18 calibrated measurements of the dataglove ($\textbf{x} \in \mathbb{R}^{18}$) are mapped to the DoA of the robotic hand ($\textbf{y} \in \mathbb{R}^{5}$) via the transformation matrix in Eq.(\ref{eq:matrix_AT}). The DoA correspond to the movement of the five fingers: $y_1$ , thumb rotation; $y_2$, thumb flexion; $y_3$, index flexion; $y_4$, middle flexion; $y_5$, ring/little finger flexion. The ring and little finger are controlled together due to a mechanical coupling in the Azzurra robotic hand \cite{krasoulis2019effect}. Using this model allows for a more explainable decoding performance and renders the model more directly transferable for the control of an external device. 

\begin{equation}
\textbf{y = Ax}
\label{eq:linear-mapping-equation}
\end{equation}

\begin{equation} \label{eq:matrix_AT}
A^{T} =
\begin{bmatrix}
0.639  & 0      & 0     & 0      & 0      \\
0.383  & 0      & 0     & 0      & 0      \\
0      & 1      & 0     & 0      & 0      \\
-0.639 & 0      & 0     & 0      & 0      \\
0      & 0      & 0.4   & 0      & 0      \\
0      & 0      & 0.6   & 0      & 0      \\
0      & 0      & 0     & 0.4    & 0      \\
0      & 0      & 0     & 0.6    & 0      \\
0      & 0      & 0     & 0      & 0      \\
0      & 0      & 0     & 0      & 0.1667 \\
0      & 0      & 0     & 0      & 0.3333 \\
0      & 0      & 0     & 0      & 0      \\
0      & 0      & 0     & 0      & 0.1667 \\
0      & 0      & 0     & 0      & 0.3333 \\
0      & 0      & 0     & 0      & 0      \\
0      & 0      & 0     & 0      & 0      \\
-0.19  & 0      & 0     & 0      & 0      \\
0      & 0      & 0     & 0      & 0      \\
\end{bmatrix}
\end{equation}

\vspace{10pt}

\section{Training and Hyperparameter Optimisation Details}
\label{sec:hparam}

% \subsection{KinEMbed Training Details}

% Contrastive pre-training uses AdamW~\cite{loshchilov2019decoupled} with weight decay
% $10^{-4}$, learning rate $3{\times}10^{-3}$, cosine annealing ($T_\text{max}{=}200$),
% batch size 256, and gradient clipping (norm 1.0). Training runs for up to 200 epochs
% with early stopping (patience 20) on a fixed 10\% validation split. Hyperparameters
% were selected via a 4-stage sequential grid search (108 combinations, 5 representative
% subjects, 2-fold session CV).

\subsection{KinEMbed}
Contrastive pre-training uses AdamW~\cite{loshchilov2019decoupled} with weight decay
$10^{-4}$, learning rate $3{\times}10^{-3}$, cosine annealing ($T_\text{max}{=}200$),
batch size 256, and gradient clipping (norm 1.0). Training runs for up to 200 epochs
with early stopping (patience 20) on a fixed 10\% validation split. 

Hyperparameters were selected via a principled 4-stage sequential grid search, designed to explore each axis of variation independently while fixing decisions from prior stages. This staged strategy reduces the  
  effective search space from a prohibitive full factorial (tens of thousands of combinations) to targeted combinations evaluated on 5 representative subjects (1, 3, 6, 8, 11; all able-bodied except subject 11) via 2-fold session CV, using mean 
  CV $R^2$ as the selection criterion. 

\textbf{Stage 1}: loss $\in$ \{InfoNCE, Soft-InfoNCE, VICReg\},
temperature $\tau \in \{0.05, 0.07, 0.1, 0.2, 0.5\}$, embedding dim $d \in \{4,8,16,32,64\}$.
\textbf{Winner}: InfoNCE, $\tau{=}0.2$, $d{=}16$.

The full per-loss comparison (best configuration per loss; mean 2-fold CV $R^2$ across
the five tuning subjects) is given in Table~\ref{tab:loss_ablation}. Standard InfoNCE
marginally outperformed the Soft-InfoNCE variant, with VICReg lowest. Although the
Soft-InfoNCE target was motivated by the false-negative problem for continuous regression
(Appendix~\ref{app:cpep}), it did not improve over standard InfoNCE in this setting,
suggesting the false-negative effect is empirically mild in our batch-local, per-subject
regime; a larger-scale study of this variant is warranted.

\begin{table}[h]
  \centering
  \caption{Stage 1 loss comparison: best configuration per loss (mean 2-fold CV $R^2$, 5 tuning subjects).}
  \label{tab:loss_ablation}
  \begin{tabular}{lc}
    \toprule
    Loss & CV $R^2$ \\
    \midrule
    InfoNCE (selected) & \textbf{0.634} \\
    Soft-InfoNCE       & 0.629 \\
    VICReg             & 0.608 \\
    \bottomrule
  \end{tabular}
\end{table}

\textbf{Stage 2}: architecture variant $\in$ \{standard, projection head,
residual, asymmetric, momentum EMA\}. \textbf{Winner}: projection head.

Table~\ref{tab:arch_ablation} reports the architecture-variant comparison. The projection
head gave only a marginal improvement over the plain (standard) encoder, indicating that
most of KinEMbed's benefit derives from the cross-modal contrastive objective rather than
the projection-head decoupling itself; the momentum-EMA variant was markedly unstable.

\begin{table}[h]
  \centering
  \caption{Stage 2 architecture-variant comparison (mean 2-fold CV $R^2$, 5 tuning subjects).}
  \label{tab:arch_ablation}
  \begin{tabular}{lc}
    \toprule
    Architecture variant & CV $R^2$ \\
    \midrule
    Projection head (selected) & \textbf{0.636} \\
    Standard                   & 0.634 \\
    Asymmetric                 & 0.629 \\
    Residual                   & 0.615 \\
    Momentum EMA               & 0.396 \\
    \bottomrule
  \end{tabular}
\end{table}

\textbf{Stage 3}: EMG hidden dims $\in \{[128,64],[256,128],[512,256,128],
[512,256,128,64]\}$, kinematic hidden dims $\in \{[32],[64,32],[128,64]\}$.
\textbf{Winner}: EMG $[256,128]$, kin $[64,32]$.

\textbf{Stage 4}: learning rate $\in \{10^{-4}, 5{\times}10^{-4},
10^{-3}, 3{\times}10^{-3}\}$, batch size $\in \{128,256,512\}$,
dropout $\in \{0.0, 0.1, 0.2\}$.
\textbf{Winner}: $\eta{=}3{\times}10^{-3}$, batch 256, dropout 0.1.

\subsection{ARIMA}

  As a time-series baseline, we evaluated an autoregressive integrated moving                                                                                                                                         
  average (ARIMA) model applied independently to each degree of actuation (DoA).                                                                                                                                         
  Unlike the EMG-feature-based methods, ARIMA operates solely on the DoA                                                                                                                                              
  time series, exploiting temporal autocorrelation in the kinematic signal                                                                                                                                            
  rather than instantaneous EMG observations.                                                                                                                                                                         
      
  Model orders $(p, d, q)$ were selected per-DoA and per-subject using the                                                                                                                                            
  \texttt{auto\_arima} procedure from the \texttt{pmdarima} library                                                                                                                                                   
  \cite{smith2017pmdarima}, with the Bayesian Information Criterion (BIC) as                                                                                                                                          
  the selection objective. BIC was preferred over AIC as its stronger                                                                                                                                                 
  complexity penalty yields parsimonious models that generalise well from                                                                                                                                             
  limited data. A stepwise search \cite{hyndman2008automatic} was used in                                                                                                                                             
  place of an exhaustive grid to reduce computation. Seasonal components were                                                                                                                                         
  not modelled, as the DoA signal has no fixed periodic structure.                                                                                                                                                    
                      
  Order selection was performed on the final 20\% of the second training                                                                                                                                              
  session (dataset~2), a single contiguous recording that is temporally                                                                                                                                               
  adjacent to the held-out test session (dataset~3). This segment provides                                                                                                                                            
  a representative characterisation of the signal's local autocorrelation                                                                                                                                             
  structure; BIC-based order selection is known to stabilise with a few                                                                                                                                               
  hundred observations \cite{hannan1979determination}.                                                                                                                                                                
                                                                                                          
  Evaluation followed the same protocol as all other methods: models were                                                                                                                                             
  applied to dataset~3. 

\subsection{CEBRA}
\label{app_cebra_hyperparams}
CEBRA hyperparameters were tuned via Optuna Tree-structured Parzen Estimator                                                                                                                                                                                             
  (TPE) search~\cite{akiba2019optuna}, run independently for each subject using                                                                                                                                                                                               
  2-fold session cross-validation with mean $R^2$ as the objective.                                                                                                                                                                                                           
  \textit{Up to 50 trials per subject} were evaluated over a categorical search space spanning                                                                                                                                                                                         
  training iterations $\in \{1000, 3000, 5000, 10000\}$, batch size $\in \{256, 512,                                                                                                                                                                                          
  1024\}$, learning rate $\in \{10^{-4}, 5{\times}10^{-4}, 10^{-3}\}$, temperature                                                                                                                                                                                            
  $\in \{0.5, 1.0, 2.0\}$, and time offset $\in \{5, 10, 20\}$ windows.                                                                                                                                                                                                       
  Canonical hyperparameters were set to the mode across subjects, yielding:                                                                                                                                                                                                   
  batch size 512, learning rate $10^{-3}$, 1000 iterations, temperature 2.0,                                                                                                                                                                                                  
  time offset 10, and embedding dimension 8 (selected in a separate dimensionality                                                                                                                                                                                            
  study); these values were fixed for all final evaluations.

\section{Additional Results}

\begin{table}[H]
  \centering
  \caption{Per-DoA $R^2$ on the held-out test set, LD subjects only ($n$=2). Values: mean $\pm$ std.}
  \label{tab:per_doa_r2_amputee}
  \resizebox{\columnwidth}{!}{%
  \begin{tabular}{lrrrrrr}
    \toprule
    Method & \multicolumn{1}{c}{Thumb rotation} & \multicolumn{1}{c}{Thumb flexion} & \multicolumn{1}{c}{Index flexion} & \multicolumn{1}{c}{Middle flexion} & \multicolumn{1}{c}{Ring/little flexion} & Mean \\
    \midrule
    ARIMA & $-0.003 \pm 0.003$ & $-0.002 \pm 0.003$ & $-0.028 \pm 0.014$ & $-0.019 \pm 0.021$ & $-0.255 \pm 0.342$ & $-0.062 \pm 0.074$ \\
    PCA & $0.618 \pm 0.114$ & $0.436 \pm 0.023$ & $0.598 \pm 0.162$ & $0.549 \pm 0.299$ & $0.599 \pm 0.249$ & $0.560 \pm 0.170$ \\
    PLS & \textbf{$0.659 \pm 0.129$} & \textbf{$0.491 \pm 0.030$} & $0.554 \pm 0.248$ & $0.517 \pm 0.344$ & $0.607 \pm 0.255$ & $0.566 \pm 0.201$ \\
    AE-Recon & $0.602 \pm 0.143$ & $0.401 \pm 0.006$ & $0.593 \pm 0.186$ & $0.569 \pm 0.253$ & $0.625 \pm 0.207$ & $0.558 \pm 0.159$ \\
    AE-Super & $0.588 \pm 0.170$ & $0.326 \pm 0.130$ & $0.538 \pm 0.256$ & $0.545 \pm 0.312$ & $0.619 \pm 0.236$ & $0.523 \pm 0.221$ \\
    CEBRA & $0.621 \pm 0.227$ & $0.360 \pm 0.120$ & \textbf{$0.620 \pm 0.185$} & \textbf{$0.617 \pm 0.184$} & \textbf{$0.630 \pm 0.188$} & \textbf{$0.570 \pm 0.181$} \\
    \midrule
    \textbf{KinEMbed} & $0.649 \pm 0.164$ & $0.416 \pm 0.106$ & $0.585 \pm 0.178$ & $0.576 \pm 0.286$ & $0.615 \pm 0.221$ & $0.568 \pm 0.191$ \\
    \bottomrule
  \end{tabular}
  }%
\end{table}

\section{Relationship to CEBRA}
  \label{sec:cebra_comparison}
                                                                                            
  CEBRA~\cite{schneider2023learnable} is a contrastive representation learning                                                                                                                                        
  framework designed to produce structured latent embeddings of neural or                                                                                                                                             
  physiological recordings by leveraging a continuous auxiliary behavioural                                                                                                                                           
  variable (or operating in a self-supervised fashion using temporal proximity). It has shown strong performance for neural data, and is thus a relevant baseline for our work. Although                                                                                                                                             
  KinEMbed shares the broad goal of learning an auxiliary-modality-structured                                                                                                                                               
  embedding of a primary physiological signal, the two methods differ in architecture, supervisory                                                                                                                                               
  mechanism, and the geometric properties of the resulting embedding space.

  \paragraph{Architecture.}                                                                                                                                                                                           
  CEBRA employs a \emph{single-encoder} design: only the primary signal                                                                                                                                       
  is passed through a parametric network, a temporal convolutional network (TCN)                                                                                                                                      
  with learnable temperature~\cite{schneider2023learnable}. Kinematics serve as a \emph{sampling oracle} that defines which                                                                                                                                      
  pairs of EMG windows should be close in the embedding space. KinEMbed instead                                                                                                                                       
  uses a \emph{dual-encoder} design: an MLP encoder processes EMG windows and a                                                                                                                                       
  separate, shallower MLP encoder processes the corresponding DoA vectors.                                                                                                                                            
  Both encoders are trained jointly, and their outputs are aligned in a shared                                                                                                                                        
  embedding space.                                                                                                                                                                              
                                                                                                            \paragraph{Positive pair construction.}
  CEBRA constructs positive pairs via a global kinematic nearest-neighbour                                                                                                                                                                                                    
  lookup: for each anchor EMG window, it searches the \emph{entire} dataset                                                                                                                                                                                                   
  for the recording whose simultaneous kinematics most closely matches a sampled kinematic target state~\cite{schneider2023learnable}.                                                                                                                                                                                                      
  KinEMbed uses a simpler synchronous scheme: the EMG window at time $t$                                                                                                                                                                                                      
  is paired directly with the simultaneously recorded DoA vector $y_t$,                                                                                                                                                                                                       
  with all other pairs in the mini-batch serving as negatives.                                                                             

  \section{Relationship to CPEP}                                                                                                                                                                                      
  \label{app:cpep}                                                                                                                                                                                                    
                                                                                                                                                                                                                      
  A concurrent line of work, CPEP~\cite{cui2025cpep}, applies cross-modal                                                                                                                                       
  contrastive learning between surface EMG and hand pose in a broadly similar                                                                                                                                         
  spirit to KinEMbed. Both methods train a dual-encoder architecture using                                                                                                                                            
  synchronised EMG--kinematic pairs and discard the kinematic encoder at                                                                                                                                              
  inference. Despite this surface similarity, the two methods differ                                                                                                                                                  
  fundamentally in task formulation, architectural design, loss construction,                                                                                                                                         
  and intended clinical context. We detail these differences below.                                                                                                                                                   
                                                                                                                                                                                                                      
  \paragraph{Task formulation.}                                                                                                                                                                                       
  The most consequential difference is the prediction target.                                                                                                                                                         
  KinEMbed is designed for \emph{continuous regression}: the decoder produces                                                                                                                                         
  real-valued estimates of five joint angles simultaneously, evaluated by                                                                                                                                             
  coefficient of determination ($R^2$) on a held-out recording session.                                                                                                                                               
  CPEP targets \emph{discrete gesture classification}.                                                                                                                                              
This distinction is not superficial. \textit{Continuous kinematic regression enables direct, proportional control}, which is essential for clinical settings as well as fluid interaction in virtual and augmented reality; discrete gesture recognition is a coarse proxy.                                                                                                                                          
  Every subsequent design choice in each method follows from this task                                                                                                                                                
  difference.                                                                                                                                                                                                         
                                                                                                                                                                                                                      
  \paragraph{The false-negative problem in regression.}                                                                                                                                                               
  Because KinEMbed targets a continuous output space, the standard InfoNCE                                                                                                                                            
  objective introduces a structural problem that does not arise in                                                                                                                                                    
  classification: two windows recorded at different times but with nearly                                                                                                                                             
  identical joint angles are treated as hard negatives and actively pushed                                                                                                                                            
  apart in the embedding space, despite representing the same kinematic state.                                                                                                                                        
  Although not used for final evaluation, we propose to address this with a \emph{Soft-InfoNCE} variant that replaces one-hot                                                                                                                                            
  positive targets with a Gaussian kernel over pairwise DoA distances,                                                                                                                                                
  weighting each negative by its kinematic proximity to the anchor and using                                                                                                                                          
  the median pairwise distance as a parameter-free bandwidth. Future work should explore the value of this loss further.                                                                                                                                                        
  CPEP does not address this problem, nor does it need to: in a discrete                                                                                                                                              
  classification setting, two windows from different gesture classes are                                                                                                                                              
  genuine negatives by construction.                                                                                                                                                                                  
                                                                                                                                                                                                                      
  \paragraph{Architecture and feature representation.}                                                                                                                                                                
  KinEMbed uses lightweight MLP encoders operating on 256-dimensional                                                                                                                                                  EMG feature vectors (time-domain and spectral statistics per                                                                                                                                           
  channel), producing 16-dimensional embeddings. A SimCLR-style projection                                                                                                                                            
  head is appended during contrastive training and discarded at inference,                                                                                                                                            
  decoupling the contrastive geometry from the downstream regression space.                                                                                                                                           
  The downstream decoder is a Temporal Convolutional Network (TCN) trained on                                                                                                                                         
  frozen embeddings to regress the five joint angles.                                                                                                                                                                 
  CPEP uses Transformer encoders (four layers, $d{=}256$) operating on raw                                                                                                                                            
  2-second EMG waveforms pre-trained via Masked Autoencoder                                                                                                                                                           
  (MAE)~\cite{he2022masked}, producing 256-dimensional embeddings. The                                                                                                                                                
  downstream task being classification, no regression decoder is trained.                                                                                                                                                                                    
                                                                                                                                       \paragraph{Dataset scale and subject population.}
KinEMbed is evaluated on NinaPro DB8~\cite{atzori2014electromyography}, comprising
11 subjects (9 able-bodied, 2 with limb difference) across three recording
sessions each. The inclusion of limb-different participants introduces
substantial variability in neuromuscular structure and EMG signal characteristics,
providing a challenging testbed for representation robustness under
distribution shift. CPEP is evaluated on the emg2pose
dataset~\cite{salter2024emg2pose}, comprising 193 participants and
approximately 370 hours of recording, and focuses entirely on able-bodied
users. \textit{The scale difference reflects different scientific goals}: CPEP
demonstrates large-scale pre-training and zero-shot transfer across users and
unseen gestures; KinEMbed demonstrates strong potential for contrastive kinematic alignment in the low-data, per-subject regime characteristic of EMG-based
control.                                                                               
  \paragraph{Evaluation protocol.}                                                                                                                                                                                    
  KinEMbed uses a strict cross-session evaluation protocol: models are trained                                                                                                                                        
  on sessions 1 and 2 and evaluated on the independently recorded session 3,                                                                                                                                          
  with no overlap. This directly tests the robustness of the embedding to                                                                                                                                             
  electrode shift and inter-session variability -- the primary failure mode for                                                                                                                                        
  clinical EMG decoders. CPEP evaluates cross-user generalisation and zero-shot                                                                                                                                       
  transfer to held-out gesture classes, which is the relevant generalisation                                                                                                                                          
  axis for a large-scale consumer application.                                                                                                                                                                        
                                                                                                                                                                                                                      
  \paragraph{Summary.}                                                                                                                                                                                                
  KinEMbed and CPEP share the cross-modal contrastive pre-training paradigm                                                                                                                                           
  but diverge in every dimension that matters for their respective applications.                                                                                                                                      
  KinEMbed is a \textit{continuous regression framework},                                                                                                                                      
  operating in the low-data regime with limb difference subjects and evaluated by joint                                                                                                                                       
  angle prediction accuracy. CPEP is a discrete\textit{ classification} method for                                                                                                                                             
  large-scale consumer gesture recognition, operating with orders of magnitude                                                                                                                                        
  more data and evaluated by gesture identification accuracy. The surface                                                                                                                                             
  resemblance in training objective reflects the generality of the cross-modal                                                                                                                                        
  contrastive framework; the differences in task, loss, architecture, data, and                                                                                                                                       
  evaluation reflect the distinct requirements of the two applications.   

\end{document}